\documentclass[11pt,a4paper]{article}

\usepackage{arxiv} 

\usepackage[T1]{fontenc}
\usepackage[utf8]{inputenc}
\usepackage{authblk}

\author[1]{Alexander Partin$^*$} 
\author[1]{Priyanka Vasanthakumari$^*$}
\author[1]{Oleksandr Narykov}
\author[1]{Andreas Wilke}
\author[2]{Natasha Koussa}
\author[2]{Sara E. Jones}
\author[1]{Yitan Zhu}
\author[1]{Jamie C. Overbeek}
\author[1]{Rajeev Jain}
\author[3]{Gayara Demini Fernando}
\author[4]{Cesar Sanchez-Villalobos}
\author[5]{Cristina Garcia-Cardona}
\author[5]{Jamaludin Mohd-Yusof}
\author[1]{Nicholas Chia}
\author[1]{Justin M. Wozniak}
\author[3]{Souparno Ghosh}
\author[4]{Ranadip Pal}
\author[1]{Thomas S. Brettin}
\author[2]{M. Ryan Weil}
\author[1,6]{Rick L. Stevens}

\affil[1]{Division of Data Science and Learning, Argonne National Laboratory, Lemont, IL, USA}
\affil[2]{Frederick National Laboratory for Cancer Research, Cancer Data Science Initiatives, Cancer Research Technology Program, Frederick, MD, USA}
\affil[3]{Department of Statistics, University of Nebraska–Lincoln, Lincoln, NE, USA}
\affil[4]{Department of Electrical \& Computer Engineering, Texas Tech University, Lubbock, TX, USA}
\affil[5]{Division of Computer, Computational and Statistical Sciences, Los Alamos National Laboratory, Los Alamos, NM, USA}
\affil[6]{Department of Computer Science, The University of Chicago, Chicago, IL, USA}

\title{Benchmarking community drug response prediction models: datasets, models, tools, and metrics for cross-dataset generalization analysis} %

\usepackage[utf8]{inputenc} 
\usepackage[T1]{fontenc}    
\usepackage{hyperref}       
\usepackage{url}            
\usepackage{booktabs}       
\usepackage{amsfonts}       
\usepackage{nicefrac}       
\usepackage{microtype}      
\usepackage{caption}
\usepackage{subcaption}
\usepackage{comment}
\usepackage{lipsum}
\usepackage[dvipsnames]{xcolor}
\usepackage{multirow}
\usepackage{soul}

\usepackage{graphicx}
\usepackage{afterpage}
\usepackage{array}
\usepackage{booktabs}
\usepackage{amssymb}
\usepackage{amsbsy}
\usepackage{amsmath}
\usepackage{longtable}

\usepackage{tabularx}    

\graphicspath{ {./figures/} }

\begin{document}
\maketitle

\def\thefootnote{*}
\footnotetext{These authors contributed equally to this work. Please cite as \textit{A. Partin and P. Vasanthakumari et al}. Correspondence: apartin@anl.gov}

\begin{abstract}

Deep learning (DL) and machine learning (ML) models have shown promise in drug response prediction (DRP), yet their ability to generalize across datasets remains an open question, raising concerns about their real-world applicability. Due to the lack of standardized benchmarking approaches, model evaluations and comparisons often rely on inconsistent datasets and evaluation criteria, making it difficult to assess true predictive capabilities.
%
%
In this work, we introduce a benchmarking framework for evaluating cross-dataset prediction generalization in DRP models.
Our framework incorporates five publicly available drug screening datasets, six standardized DRP models, and a scalable workflow for systematic evaluation.
To assess model generalization, we introduce a set of evaluation metrics that quantify both absolute performance (e.g., predictive accuracy across datasets) and relative performance (e.g., performance drop compared to within-dataset results), enabling a more comprehensive assessment of model transferability. 
%
%
Our results reveal substantial performance drops when models are tested on unseen datasets, underscoring the importance of rigorous generalization assessments.
While several models demonstrate relatively strong cross-dataset generalization, no single model consistently outperforms across all datasets. 
Furthermore, we identify CTRPv2 as the most effective source dataset for training, yielding higher generalization scores across target datasets.
%
%
By sharing this standardized evaluation framework with the community, our study aims to establish a rigorous foundation for model comparison, and accelerate the development of robust DRP models for real-world applications.

\end{abstract}


\section{Introduction} \label{sec:intro}

Predictive modeling has become an integral tool in cancer research, offering new capabilities for improving clinical outcomes \cite{singhalHallmarksPredictiveOncology2025, senftPrecisionOncologyRoad2017}.
Anti-cancer drug response prediction (DRP) models provide an in-silico approach to evaluating the potential effects of drugs on cancer, leveraging artificial intelligence (AI) techniques such as deep learning (DL) and classical machine learning (ML) \cite{partinDeepLearningMethods2023, ballesterArtificialIntelligenceDrug2021, adamMachineLearningApproaches2020}.
The increasing availability of high-throughput genomic profiling and drug screening data has enabled the rapid development of these models.
However, despite the growing number of DRP models, it remains unclear how well their predictive performance translates to unseen datasets and experimental conditions, limiting their potential for broader real-world applicability \cite{singhalHallmarksPredictiveOncology2025, partinDeepLearningMethods2023, sharifi-noghabiDrugSensitivityPrediction2021}.



While the ultimate goal is to accurately predict drug responses in patients, in-vitro cell line drug screenings remain the most abundant source of data and, therefore, the primary resource for developing and evaluating DRP models.
However, while many published models achieve high predictive accuracy within a single cell line dataset, their performance deteriorates when applied to more complex biological systems such as organoids, patient-derived xenografts (PDX), or patients samples.
Therefore, an intermediate step in complexity involves assessing prediction generalization across different cell line datasets, a practice that has recently become widely common \cite{sharifi-noghabiDrugSensitivityPrediction2021}.
Demonstrating robust cross-dataset generalization goes beyond simple cross-validation within a single dataset and positions a model as a promising candidate for more complex transfer tasks.


Despite the growing number of DRP models, the lack of standardized frameworks for scalable benchmarking presents a major challenge in identifying the most effective methods for cancer treatment prediction. Performance evaluation is often conducted with simple baselines, varied dataset compositions, inconsistent data splits, and diverse scoring metrics, making it difficult to assess model strengths, weaknesses, and key factors contributing to model performance.
To demonstrate tangible progress in the field and better understand model capabilities, DRP models should undergo systematic cross-comparisons against multiple other models, ideally demonstrating sufficient cross-dataset generalization, in a rigorous and systematic way.
Recognizing this need, collaborative community efforts have emerged to establish guidelines and best practices for improving model development and evaluation \cite{singhalHallmarksPredictiveOncology2025, sharifi-noghabiDrugSensitivityPrediction2021, partinDeepLearningMethods2023}.





To address the pressing need for standardized benchmarking in DRP, this study, conducted as part of IMPROVE \cite{weil15PB003IMPROVE2024, overbeekAssessingReusabilityDeep2024}
project, aims to establish a framework for systematic evaluation and comparison of DL models, applicable to various scientific domains.
We present simple benchmarking principles centered around four key aspects:
1) benchmark datasets,
2) standardized models,
3) software tools and protocols, and
4) evaluation workflows.
These principles, which form the core contributions of this study, are designed to facilitate systematic model benchmarking. Specifically, we constructed a comprehensive benchmark dataset (Section \ref{sec:dataset}) comprising drug response data from five publicly available drug screening studies (Table \ref{tab:datasets}), along with drug and omics feature representations and pre-computed data splits to ensure consistency across evaluations.
We also designed a unified code structure promoting modular design and developed \textit{improvelib}, a lightweight Python package that standardizes preprocessing, training, and evaluation, ensuring consistent model execution and enhancing reproducibility. 
Furthermore, we selected five DL-based DRP models and one ML model built with LightGBM \cite{keLightGBMHighlyEfficient2017} (Table \ref{tab:list_of_models}) and adjusted their code to conform to the required modular structure. 
Finally, we implemented a scalable workflow to conduct large-scale cross-dataset analysis, applying it to the selected models and benchmark dataset for comprehensive training and evaluation.

\section{Materials and Methods} \label{sec:methods}

\begin{figure}[h]
    \centering
    \includegraphics[width=0.99\textwidth]{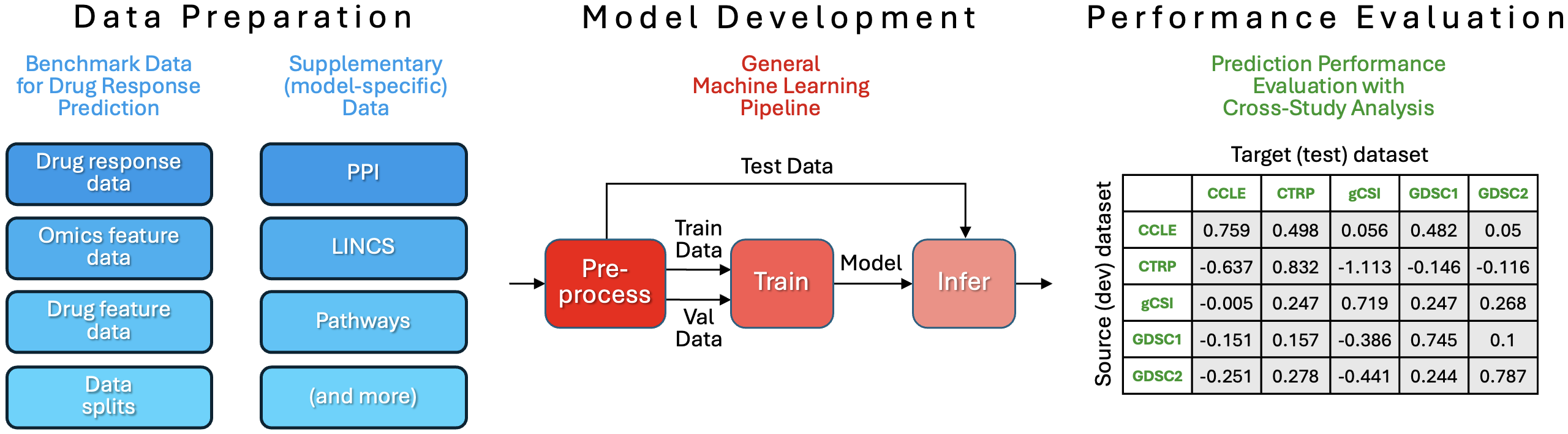}
    \caption{Basic components in the development of drug response prediction (DRP) models. The process consists of three main stages: (1) Data Preparation: including benchmark data components (e.g., drug response, omics, and drug feature data) and supplementary model-specific data; (2) Model Development: a general ML pipeline structured into three distinct stages—preprocessing, training, and inference; (3) Performance Evaluation: cross-dataset generalization assessment (the example $G$ matrix is shown for illustration purposes and does not represent real experimental results).
    }
    \label{fig:basic_drp_comp}
\end{figure}

The development of DRP models, much like any supervised learning model, typically includes three core components: data preparation, model development, and performance analysis \cite{partinDeepLearningMethods2023}. The specifics of each component are often contingent upon the application, performance evaluation scheme, required rigor, and scale of the analysis. Within this general framework, we outline the details of the analysis conducted in this work (Fig. \ref{fig:basic_drp_comp}).

A benchmark dataset, the product of the data preparation step, is essential for fair analysis and a required ingredient for establishing state-of-the-art (SotA) among models. In the context of our analysis, which investigates DRP models, this dataset comprises drug response data compiled from multiple drug screening studies, omics and drug feature data, and pre-computed data splits to ensure consistent train, validation, and test sets.

Model development, in this context, refers to an ML learning pipeline comprising data preprocessing, model training, and inference. These steps take the output of data preparation (benchmark data and potentially additional model-specific data), transform it into suitable model input data, train the model, and generate predictions on a hold-out set through inference.

Finally, the evaluation scheme can encompass a wide array of approaches, ranging from basic cross-validation or hold-out set analysis on a single dataset to more complex schemes. These may include cross-dataset generalization across studies from the same \cite{zhuEnsembleTransferLearning2020, parkSuperFELTSupervisedFeature2021, xiaCrossstudyAnalysisDrug2022} or different biological media \cite{sharifi-noghabiAITLAdversarialInductive2020, maFewshotLearningCreates2021, peresdasilvaTUGDATaskUncertainty2021, sharifi-noghabiOutofdistributionGeneralizationLabelled2021}, data scaling analysis via learning curves \cite{liuImprovingPredictionPhenotypic2019, partinLearningCurvesDrug2021, xiaCrossstudyAnalysisDrug2022, bransonComparisonMultipleModalities2024}, and interpretability assessments \cite{dengPathwayGuidedDeepNeural2020, liInterpretableDeepLearning2023}.
In this work, a workflow for conducting scalable cross-dataset generalization analysis with the benchmark dataset and six DRP models is presented.

The rest of the Methods section details our approach, demonstrated with six prediction models and the benchmark dataset. We describe the benchmark dataset's characteristics (Section \ref{sec:dataset}), the models and the model selection criteria (Section \ref{sec:models}), the methodology for standardizing models to ensure consistent evaluation (Section \ref{sec:model_standard}), and the cross-dataset evaluation scheme used (Section \ref{sec:csa}).

\subsection{Benchmark dataset} \label{sec:dataset}

The benchmark dataset consists of three main components: response data, cancer features, and drug features. These components include data types commonly used by pan-cancer and multi-drug models for predicting treatment response \cite{partinDeepLearningMethods2023}. Unique cancer and drug identifiers link the components, allowing integration into structured formats for model training and evaluation.

\begin{figure}[h]
    \centering
    \includegraphics[width=0.99\textwidth]{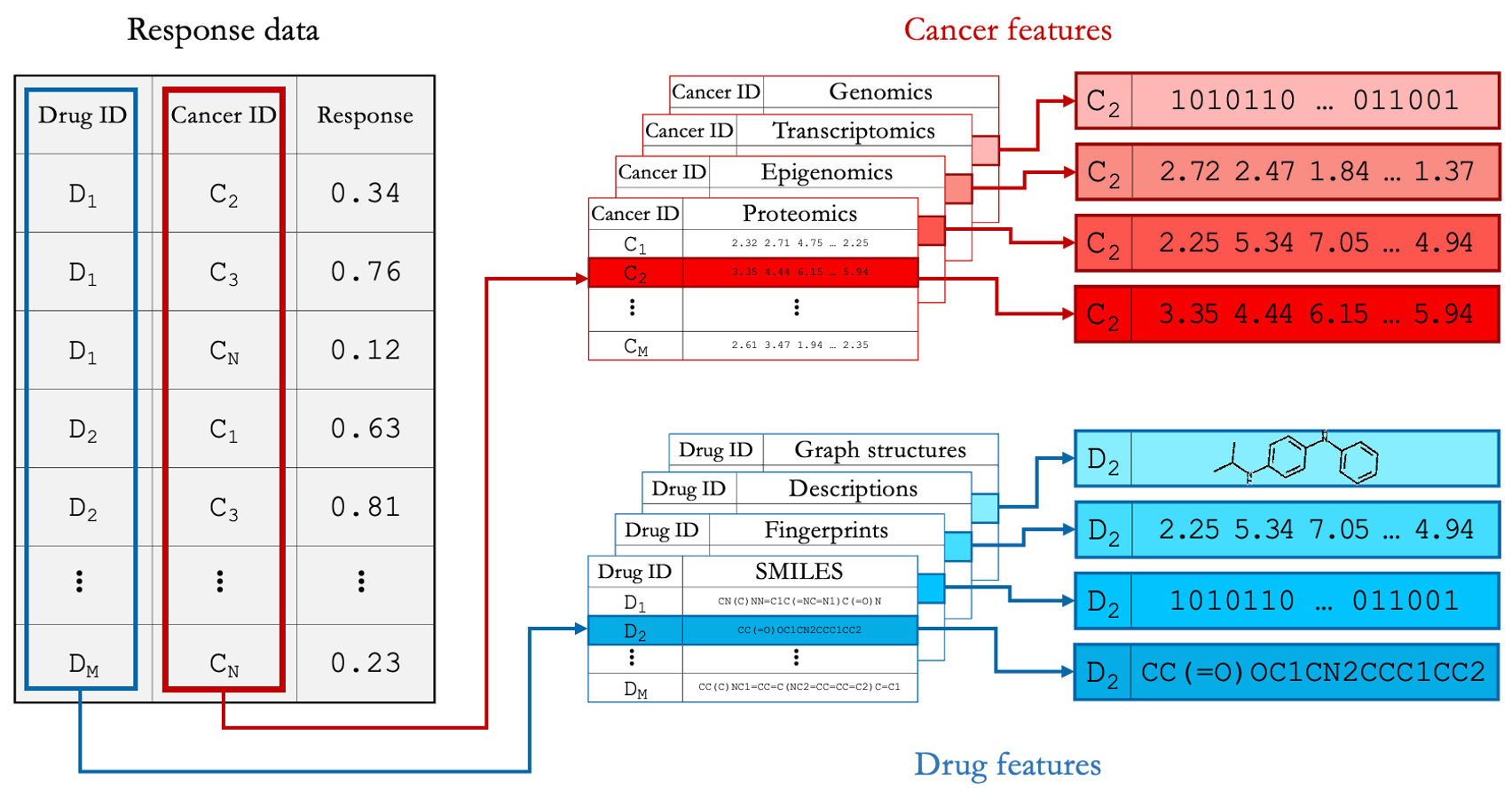}
    \caption{The main data components in the drug response benchmark dataset and their integration through shared cell and drug identifiers.}
    \label{fig:benchmark_data}
\end{figure}





\subsubsection{Drug response data} \label{sec:drp_data}

The drug response data was gathered from five public drug screening studies: Cancer Cell Line Encyclopedia (CCLE) \cite{barretinaCancerCellLine2012}, Cancer Therapeutics Response Portal (CTRPv2) \cite{seashore-ludlowHarnessingConnectivityLargescale2015}, Genentech Cell Line Screening Initiative (gCSI) \cite{havertyReproduciblePharmacogenomicProfiling2016}, and Genomics of Drug Sensitivity in Cancer project, which includes GDSCv1 and GDSCv2 datasets \cite{yangGenomicsDrugSensitivity2013}.
The response in cell lines was quantified by measuring cell viability across multiple drug doses. We fitted each cell-drug pair's dose-response data to a three-parameter Hill-Slope curve, excluding pairs with R² < 0.3 to ensure data quality. The area under the curve (AUC) was calculated over a dose range of $[10^{-10}$M, $10^{-4}$M$]$ and normalized to [0, 1], with lower AUC values indicating stronger response (higher growth inhibition).
Each cell-drug pair AUC value represents a single drug response sample. Table \ref{tab:datasets} summarizes the number of unique drugs, cell lines, and total AUC responses in each dataset, constituting the benchmark data.

\begin{table}[h]
\caption{Summary of the number of unique drugs, cell lines, and AUC response across drug screening datasets.}
   \centering
   \begin{tabular}{l | rrrrr}
       \toprule
       Dataset & Drugs & Cell Lines & Responses\\
       \midrule
       CCLE & 24\hspace{5pt} & 411\hspace{10pt} & 9,519\hspace{2pt}\\
       CTRPv2 & 494\hspace{5pt} & 720\hspace{10pt} & 286,665\hspace{2pt}\\
       gCSI & 16\hspace{5pt} & 312\hspace{10pt} & 4,941\hspace{2pt}\\
       GDSCv1 & 294\hspace{5pt} & 546\hspace{10pt} & 171,940\hspace{2pt}\\
       GDSCv2 & 168\hspace{5pt} & 470\hspace{10pt} & 114,644\hspace{2pt}\\
       \bottomrule
   \end{tabular}
  \label{tab:datasets}
\end{table}

\subsubsection{Multiomics data} \label{sec:omics_data}
The multiomics data used for cell line representation were sourced from the Dependency Map (DepMap) portal of CCLE version 22Q2 \footnote{https://depmap.org/portal} \cite{tsherniakDefiningCancerDependency2017}. This dataset includes gene expressions, DNA mutations, DNA methylation, gene copy numbers, protein expressions (RPPA), and miRNA expressions. 
The gene expression data, derived from RNA-seq and log2-transformed with a pseudo count of 1, covers 1,007 cell lines across 30,805 genes. 
The mutation data in long format provides details on somatic point mutations for 1,024 cell lines, including the mutated gene, variant classification, genome change, protein change, and chromosomal positions. Additionally, two derivative datasets were curated: binary mutation data, which identifies the presence of mutations in a gene within a cell line, and mutation count data, representing the number of mutations per gene in each cell line. 
The DNA methylation data contains methylation levels at transcription start sites (TSS) for 824 cell lines. 
The copy number data includes log2-transformed gene-level copy number values for 1,018 cell lines across 25,331 genes. 
The RPPA data reports protein expression levels for 789 cell lines mapped to 214 target genes through antibody-based proteomics. 
The miRNA data includes expression levels for 820 cell lines across 734 miRNAs.

\subsubsection{Drug representation data} \label{sec:drugs_data}

In addition to drug response and omics data, the benchmark includes drug metadata and three types of drug representations: SMILES, fingerprints, and descriptors.
Although SMILES are rarely used in their raw string format in DRP models, they remain a popular notation for describing molecules. SMILES often serve as an intermediate step for generating other representations, such as fingerprints, molecular descriptors, and graph structures. Both raw and canonicalized SMILES are provided in the dataset.
Fingerprints (FPs) and molecular descriptors are perhaps the two most common feature types for representing drugs in DRP papers \cite{partinDeepLearningMethods2023}. Unlike SMILES, where string length varies for different molecules, FPs and descriptors allow for a consistent number of features across all drugs in a dataset. This fixed feature dimensionality makes them easier to use with prediction models.
We used the open-source package RDKit \cite{rdkit} to generate 512-bit Extended Connectivity Fingerprints (ECFPs) via the Morgan algorithm.
Descriptors were calculated using the open-source Mordred package \cite{moriwakiMordredMolecularDescriptor2018}. The full descriptor set comprises 1,826 continuous and discrete features, including both 2D and 3D molecular structure descriptors. Since most 3D descriptors resulted in invalid (NaN) values for the majority of compounds, we retained only the 2D descriptors, providing a total of 1,613 drug features.

\subsubsection{Data partitions} \label{sec:splits_data}


The final component of the benchmark data comprises multiple sets of disjoint data splits, used to ensure consistent training, validation, and test data utilized across all models, thereby enhancing the rigor of the analysis.
For each of the five datasets in Table \ref{tab:datasets}, we generated ten data splits using random 10-fold cross-validation, resulting in ten sets of training ($T$), validation ($V$), and test ($E$) samples per dataset, represented as $\{T, V, E\}_{n=1}^{N}$ where $N=10$.
Each set $\{T, V, E\}$ maintains a size proportion of (0.8, 0.1, 0.1) relative to the total number of drug response samples in a dataset.
The splits are stored as text files containing row indices corresponding to the drug response data, allowing for easy retrieval of specific subsets.

\subsection{Models} \label{sec:models}
We previously compiled a comprehensive list of more than 60 papers (as of Aug 2022) proposing DL-based models for monotherapy DRP \cite{partinDeepLearningMethods2023}. From this list, we selected 17 models for a reusability and reproducibility study, assessing the ease of reusing and adapting these models to new datasets and contexts \cite{overbeekAssessingReusabilityDeep2024}. In the current study, we further refined this selection to six models from the reusability study and included one additional model, called UNO \cite{xiaCrossstudyAnalysisDrug2022}. Section \ref{sec:model_selection} outlines the primary selection criteria used in \cite{overbeekAssessingReusabilityDeep2024} that remain applicable here, along with additional criteria leading to the final selection of models listed in Table \ref{tab:list_of_models}.

\subsubsection{Model Selection Process} \label{sec:model_selection}
We selected a subset of models based on qualitative and empirical criteria.


Qualitative criteria:
\begin{enumerate}
    \item Model is accompanied by open-source code, with preference given to publications from 2019 onwards
    \item Deep learning model implemented in PyTorch or TensorFlow/Keras
    \item Model predicts monotherapy drug response values using an end-to-end learning approach
    \item Code repository includes clear instructions on how to use the model and reproduce results
    \item Pan-cancer and pan-drug model (i.e., learns from cancer and drug representations) 
\end{enumerate}

Empirical criteria:
\begin{enumerate}
    \item We successfully installed the computational environment
    \item We successfully executed preprocessing scripts and generate model-input data
    \item We were able to reproduce key results reported in the original publications
\end{enumerate}


\subsubsection{Selected Models} \label{selected_models}
Following the selection process outlined in \ref{sec:model_selection}, a total of five DL models were selected for this study together with LGBM, a LightGBM-based model. Table \ref{tab:list_of_models} lists these models, including the required omics and drug representations, and the unique neural network building blocks employed in each model's architecture.
The selected models demonstrate a range of design approaches, encompassing single- to multi-modal architectures and incorporating diverse neural network components such as convolutional neural networks, graph neural networks, and attention mechanisms.

\begin{table}[t]
\caption{\textbf{Overview of Models and Feature Representations}. This table summarizes the selected models, computational frameworks, and the cancer and drug feature types incorporated from the benchmark dataset. Some models utilize additional unique representations not detailed here. Abbreviations: TF – TensorFlow; GE – gene expression; Methyl – methylation; Mu – mutation; CNV – copy number variation; DD – drug descriptors; MG – molecular graph; FPs – fingerprints; SMILES – simplified molecular input line entry system; GIN – graph isomorphism network; 1D-CNN – one-dimensional convolutional neural network; RC – residual connections.
}
    \centering
    \resizebox{\columnwidth}{!}{
    \begin{tabular}{@{}clllllll@{}}
        \toprule
        & Model & Year & Framework & Cancer Omics Features & Drug Features & Architecture components\\
        \midrule
        1 & DeepCDR \cite{liuDeepCDRHybridGraph2020} & 2020 & TF-Keras & GE, Methyl, Mu & MG & Batchnorm, Dropout\\ 
        2 & GraphDRP \cite{nguyenGraphConvolutionalNetworks2021} & 2022 & PyTorch & CNV, Mu & MG & Batchnorm, Dropout, GIN (drugs), 1D-CNN (GE)\\
        3 & HiDRA \cite{jinHiDRAHierarchicalNetwork2021} & 2021 & TF-Keras & GE & FPs & Attention, Batchnorm\\ 
        4 & tCNNs \cite{liuImprovingPredictionPhenotypic2019} & 2019 & TF & CNV, Mu & SMILES (one-hot) & 1D-CNN (CNV, Mu, SMILES), Dropout\\
        5 & UNO \cite{xiaCrossstudyAnalysisDrug2022} & 2019 & TF-Keras & GE & DD, FPs & Dropout, RC\\
        6 & LGBM & & LightGBM & GE & DD & \\
\bottomrule
\end{tabular}
}
\label{tab:list_of_models}
\end{table}

\subsection{Standardizing Models} \label{sec:model_standard}


Computational workflows can be designed to conduct systematic and rigorous analyses across various models and experimental conditions.
The inconsistent code structures of different models hinder the straightforward design of such workflows.
We address this by standardizing models through several key principles.
An essential prerequisite for designing workflows is that the models' code should be structured in a modular fashion, allowing easy, upon-request invocation of specific modules consistently across models.
This approach enables workflow developers to design workflows without requiring in-depth knowledge of specific model details.

\begin{figure}[h]
    \centering
    \includegraphics[width=1.0\textwidth]{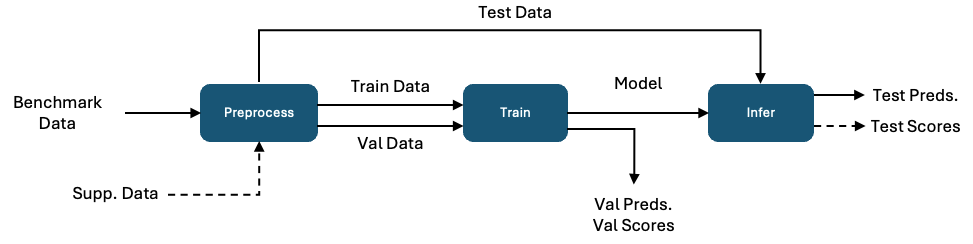}
    \caption{General machine learning (ML) pipeline. This includes preprocessing, training, and inference.  The dashed lines signify optional elements (\textit{Supp. Data} refers to data utilized by the model that is not part of the benchmark dataset).
    }
    \label{fig:gen_ml_pipeline}
\end{figure}

The first principle involves standardized code structure. Prediction models, regardless of the application, generally consist of three main stages in the ML pipeline (see Fig. \ref{fig:gen_ml_pipeline}): 1) \textbf{preprocessing} to transform benchmark data into model-input data, 2) \textbf{training} to obtain a prediction model, and 3) \textbf{inference} to compute predictions. We require the separation of these stages into distinct model scripts, allowing flexible workflow design where each stage can be called as a module.

Second, we enforce a unified interface for interacting with these scripts, ensuring consistent inputs and outputs. To create a generalizable design applicable across various scientific domains, the interface supports three sets of input parameters for each model script: 1) general \textit{improvelib} parameters, 2) application-specific parameters (where \textit{application} refers to a specific use case such as DRP), and model-specific parameters. This tiered parameter structure enhances flexibility and generalizability across diverse applications and models.

The third component, crucial for achieving model standardization, involves a set of utility tools incorporated in the \textit{improvelib} Python package. This lightweight package, with minimal standard dependencies, provides the necessary functionalities to support the three model stage scripts and the unified interface. It enables proper parameter precedence (command line, config file, defaults), generates user-friendly help messages, and facilitates logging of experimental conditions.

Figure \ref{fig:gen_ml_pipeline} demonstrate the general ML pipeline, encompassing preprocessing, training, and inference stages.
Table \ref{tab:io_scripts} details the inputs, outputs, and key functionalities provided by \textit{improvelib} for each ML stage, ensuring consistently across all standardized models.

\textbf{Preprocessing.} The preprocessing script handles benchmark data in all models we evaluate.
The Python package includes designated data loaders for various benchmark files, including cancer and drug feature data ($x$ data), drug response data ($y$ data), and data splits.
It is expected that all models use these loaders to load benchmark data files.
Some models utilize additional data types beyond what is included in the benchmark data (e.g., protein-protein interactions (PPI) \cite{jinHiDRAHierarchicalNetwork2021}).
The preprocessing interface accommodates this via an optional input argument specifying the path to such model-specific data.
The preprocessing script then utilizes this path to load the additional data and appropriately combine it with the benchmark data as required by the model.
The output of this stage consists of three model-ready datasets: training, validation, and test data, each formatted specifically for the model's framework (e.g., \textit{pt} for PyTorch, \textit{tfrecords} for TensorFlow).

\textbf{Training}. The training script uses the preprocessed training and validation data to produce a trained model, along with raw predictions and performance scores for the validation set.
To ensure convergence, all models implement early-stopping and use validation data to terminate training if validation loss does not improve for a predefined number of iterations (the model with the lowest validation loss is saved for the inference stage).
The package includes standardized methods for saving the trained model to a predefined path and store predictions and scores with consistent naming conventions. This simplifies model loading during inference and facilitates result aggregation in post-processing.

\textbf{Inference}. Finally, the inference script takes the trained model and applies it to the preprocessed test data, generating and saving test set predictions and performance scores. Similar to the validation process, test predictions and scores are stored in consistently named output files, which are then aggregated in post-processing script to produce the final results.

\begin{table}[h]
\caption{Primary inputs, outputs, and functionality of preprocessing, training, and inference scripts.}
    \centering
    \begin{tabular}{l | lll} 
        \toprule
        & Inputs & Outputs & \textit{improvelib} functionality\\
        \midrule
        Preprocessing & Benchmark data & Train data & \textit{DRPPreprocessConfig}\\
        & Supplementary data & Val data & \textit{OmicsLoader, DrugsLoader,}\\
        & & Test data & \textit{DrugResponseLoader}\\
        \midrule
        Training & Train data & Val data predictions & \textit{DRPTrainConfig,} \\
        & Val data & Val data performance scores & \textit{store\_predictions\_df,} \\
        & & Trained model & \textit{compute\_performance\_scores}\\
        \midrule
        Inference & Test data & Test data predictions & \textit{DRPInferConfig} \\
        & Trained model & Test data performance scores (optional) & \textit{store\_predictions\_df} \\
        & & & \textit{compute\_performance\_scores}\\
        \bottomrule
    \end{tabular}
    \label{tab:io_scripts}
\end{table}

\subsection{Cross-Dataset Workflow} \label{sec:csa}

In our prior works \cite{xiaCrossstudyAnalysisDrug2022, partinAbstract5380Systematic2023}, we conducted cross-study analysis to evaluate prediction generalization of models across multiple drug screening studies. However, those analyses were limited to a smaller set of models and were not designed as a systematic benchmarking framework. In this work, we extend that methodology and adopt the broader term \textit{cross-dataset generalization}, which reflects its applicability beyond biological studies.

The cross-dataset evaluation scheme assesses prediction generalization both across and within datasets. 
The \textit{source} dataset is the drug response data from a drug screening dataset used for model development, including training and validation data. 
The \textit{target} dataset is a separate set of drug response samples used exclusively for model evaluation, which may either be from the same dataset (within-dataset evaluation) or a different dataset (cross-dataset evaluation).

We utilized \( d = 5 \) datasets (see Table \ref{tab:datasets}), resulting in a \( d \)-by-\( d \) matrix \( G \) for each model. The diagonal entries of \( G \), \( g[s,s] \), represent within-dataset prediction performance, while the off-diagonal entries, \( g[s,t] \) for \( s \neq t \), represent cross-dataset performance. To compute a single entry in \( G \), the workflow executes the ML pipeline (see Fig. \ref{fig:gen_ml_pipeline}) \( N \) times for a given source-target datasets pair. Here, \( N \) is the total number of data splits generated for each drug response dataset, where \( N = 10 \) in our analysis. The scores from these \( N \) executions are averaged via

\[
g[s,t] = \frac{1}{N} \sum_{n=1}^N g[s,t,n]
\]

and each averaged score \( g[s,t] \) is assigned to the corresponding entry in \( G \) (further described in \ref{sec:metrics}).

Hyperparameter selection was limited to batch size, learning rate, and dropout rate (for models incorporating dropout layers). These values were manually adjusted based on the original model publications but were not subjected to exhaustive tuning. This approach ensured that training configurations remained aligned with prior studies while allowing for stable performance across datasets.

\begin{figure}[h]
    \centering
    \includegraphics[width=0.99\textwidth]{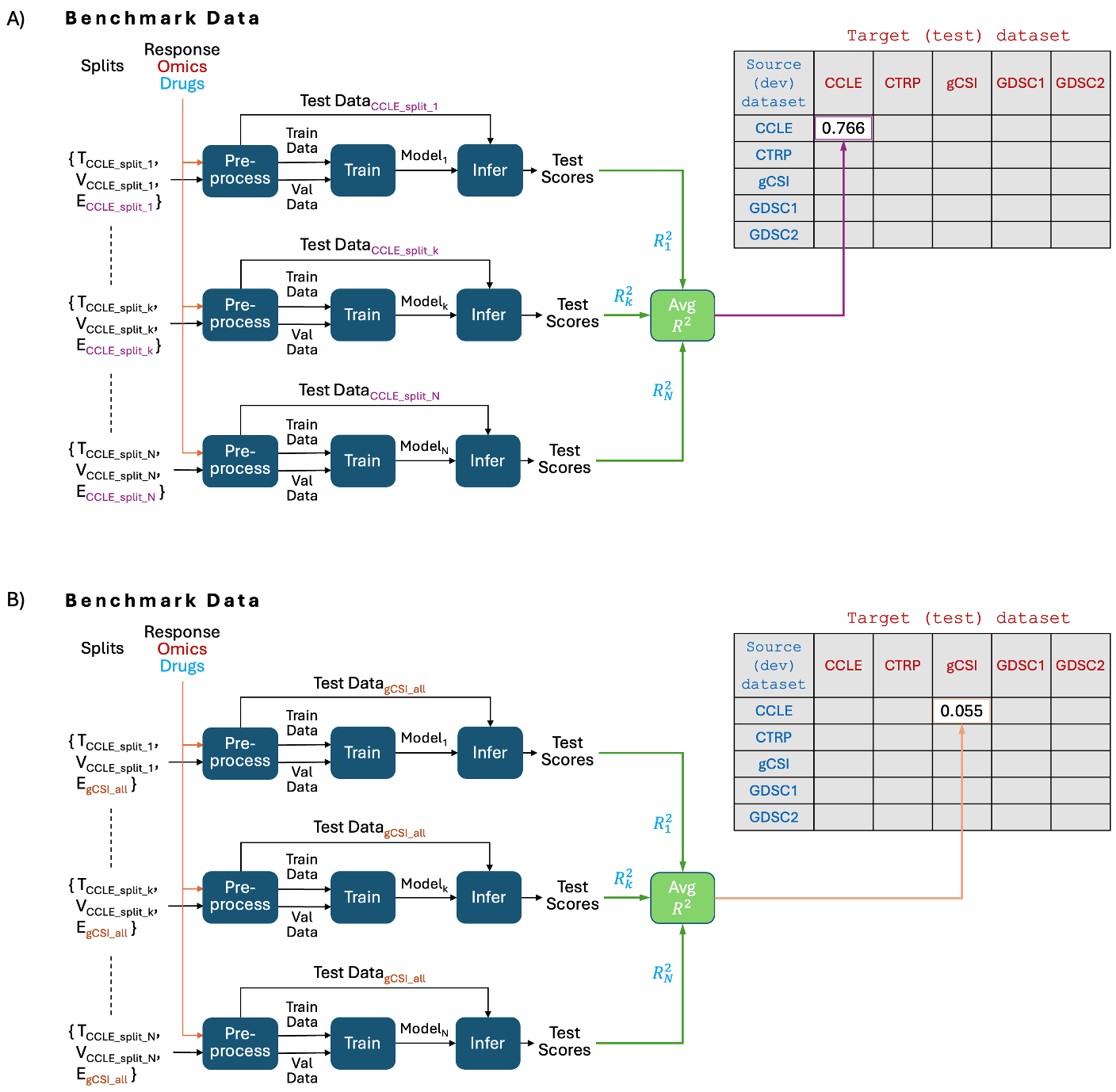}
    \caption{\textbf{Cross-Dataset Generalization}. A. Computing a prediction performance score for the \( G[CCLE, CCLE] \) matrix entry (within-study case). B. Computing a prediction performance score for the \( G[CCLE, gCSI] \) matrix entry (cross-dataset case). The parallel implementation of this workflow is implemented using Parsl.
    }
    \label{fig:csa_wf}
\end{figure}

Figures \ref{fig:csa_wf}A and \ref{fig:csa_wf}B illustrate the workflows for generating within-dataset and cross-dataset results, respectively. 
In Figure \ref{fig:csa_wf}A, training, validation, and test data originate from the same dataset (e.g., CCLE split 0), and the averaged score \( g[s,s] \) is assigned to the diagonal entry in \( G \). 
In Figure \ref{fig:csa_wf}B, training and validation data are from one dataset (e.g., CCLE split 0), while the test data is from a different dataset (e.g., gCSI). 
The averaged score \( g[s,t] \) is then assigned to the corresponding off-diagonal entry in \( G \).

It is important to note that although the response values (\( y \)) in the target dataset remain constant across ML pipeline runs in the case of Figure \ref{fig:csa_wf}B, the feature values (\( x \)) may vary due to preprocessing steps like feature scaling.  This preprocessing, often dependent on the training data, may introduce variation in the feature values of the target dataset across different pipeline runs.

As shown in Figure \ref{fig:csa_wf}, the ML pipelines for \( N \) splits can be executed in parallel. 
To facilitate parallel execution, we implemented this workflow using the Parsl parallel processing library \cite{babujiParslPervasiveParallel2019}.
Each pipeline component—preprocessing, training, and inference—was designed as a Parsl app, returning a \textit{futures} object to monitor execution progress.
For instance, once preprocessing for CCLE split 1 is complete, the corresponding \textit{futures} object triggers training for that split. 
Similarly, inference is initiated upon training completion, with its own \textit{futures} object. 
This parallelized workflow was executed for all models on an 8-GPU cluster at Argonne National Laboratory.

\subsection{Performance Metrics} \label{sec:metrics}

In our previous work \cite{xiaCrossstudyAnalysisDrug2022}, we analyzed cross-dataset results by presenting raw prediction performance scores in a \( d \)-by-\( d \) matrix, where \( d \) represents the number of datasets. Here, we extend that framework by introducing four key metrics: \( G \), \( G_a \), \( G_n \), and \( G_{na} \). These metrics provide deeper insights into the generalization capabilities of models by capturing both absolute cross-dataset performance and performance relative to the observed within-dataset performance.

\subsubsection{Basic Cross-Dataset Performance Matrix (\( G \))} \label{sec:G_matrix}

The matrix \( G \) represents the performance of a model trained on a source dataset \( s \) and evaluated on a target dataset \( t \), where \( s, t \in \{1, 2, \dots, d\} \) are the indices of the datasets included in the analysis. Each entry \( g[s,t] \) in \( G \) represents the average predictive performance of the model across \( N \) independent data splits:

\[
g[s,t] = \frac{1}{N} \sum_{n=1}^N g[s,t,n], \quad \forall \, s, t \in \{1, 2, \dots, d\}
\]

where \( g[s,t,n] \) denotes the performance score (e.g., \( R^2 \)) for the \( n \)-th split of the model trained on \( s \) and evaluated on \( t \).

In our previous work \cite{xiaCrossstudyAnalysisDrug2022}, the \( G \) matrix was introduced to illustrate cross-dataset performance, though without a structured benchmarking framework. Here, we formally establish it as part of a systematic evaluation methodology.
In the current study, we expand on this concept by ensuring consistent data splits across all evaluated models. Additionally, we report both the mean and standard deviation of performance scores for each entry \( g[s,t] \), providing a more comprehensive view of cross-dataset variability (Fig. \ref{fig:G_and_Gn_matrices}).

\subsubsection{Aggregated Cross-Dataset Performance (\( G_a \))} \label{sec:Ga_matrix}

The metric \( G_a \) aggregates the entries of \( G \) across all target datasets, excluding the within-dataset evaluation (\( s = t \)), to summarize the cross-dataset performance of a model trained on a given source dataset:

\[
g_a[s] = \frac{1}{d-1} \sum_{\substack{t=1 \\ t \neq s}}^d g[s,t], \quad \forall \, s \in \{1, 2, \dots, d\}
\]

where \( g[s,t] \) is an entry in \( G \). This aggregation provides a single summary value per source dataset, indicating how well a model trained on \( s \) generalizes on average to all other datasets. It highlights the intrinsic generalization capability of a source dataset without being influenced by within-dataset performance.

\subsubsection{Normalized Cross-Dataset Performance Matrix (\( G_n \))} \label{sec:Gn_matrix}

The normalized matrix \( G_n \) extends \( G \) by scaling each entry \( g[s,t] \) by the corresponding within-dataset performance \( g[s,s] \). This normalization provides a relative measure of generalization:

\[
g_n[s,t] = \frac{g[s,t]}{g[s,s]}, \quad \forall \, s, t \in \{1, 2, \dots, d\}
\]

The normalized performance matrix \( G_n \) contextualizes each model's performance on a target dataset \( g[s,t] \) relative to its baseline within-dataset performance \( g[s,s] \). This scaling quantifies the degree to which models trained on a source dataset \( s \) generalize to target dataset \( t \), highlighting whether their performance on \( t \) approaches or maintains the level observed within \( s \) itself.

\subsubsection{Aggregated Normalized Cross-Dataset Performance (\( G_{na} \))} \label{sec:Gna_matrix}

The metric \( G_{na} \) aggregates the normalized performance scores \( g_n[s,t] \) across all target datasets (excluding \( s = t \)) to summarize cross-dataset generalization relative to within-dataset performance:

\[
g_{na}[s] = \frac{1}{d-1} \sum_{\substack{t=1 \\ t \neq s}}^d g_n[s,t], \quad \forall \, s \in \{1, 2, \dots, d\}
\]

This metric provides a single summary score per source dataset \( s \), emphasizing how well a model generalizes across other datasets while accounting for its within-dataset performance. 

These metrics collectively offer a comprehensive framework for evaluating cross-dataset performance. While \( G \) and \( G_a \) focus on absolute cross-dataset performance, \( G_n \) and \( G_{na} \) emphasize relative performance normalized by within-dataset results. Together, they provide insights into both pairwise dataset relationships and dataset-level generalization trends, offering tools to assess the broader applicability and robustness of models.

\section{Results} \label{sec:results}

Despite leveraging advanced and diverse DL techniques (Table \ref{tab:list_of_models}), all models demonstrate substantially lower performance in cross-dataset analysis compared to cross-validation within a single dataset (Fig. \ref{fig:G_and_Gn_matrices}). This disparity underscores the inherent difficulty of achieving cross-dataset generalization in DRP and highlights the need for rigorous and systematic evaluation frameworks that closely reflect the complexities of real-world applications.

\subsection{Within-Dataset Performance} \label{sec:within_dataset_performance}



The within-dataset evaluation measures a model's predictive performance when training and testing are conducted on the same dataset. This analysis provides key insights into model performance and robustness under controlled conditions.
Tables \ref{tab:within_dataset_mean_r2} and \ref{tab:within_dataset_std_r2} summarize the mean and standard deviation of \( R^2 \) scores across splits, respectively, reflecting on performance generalization and stability of these models.

\begin{figure}[h]
    \centering
    \includegraphics[width=1.0\textwidth]{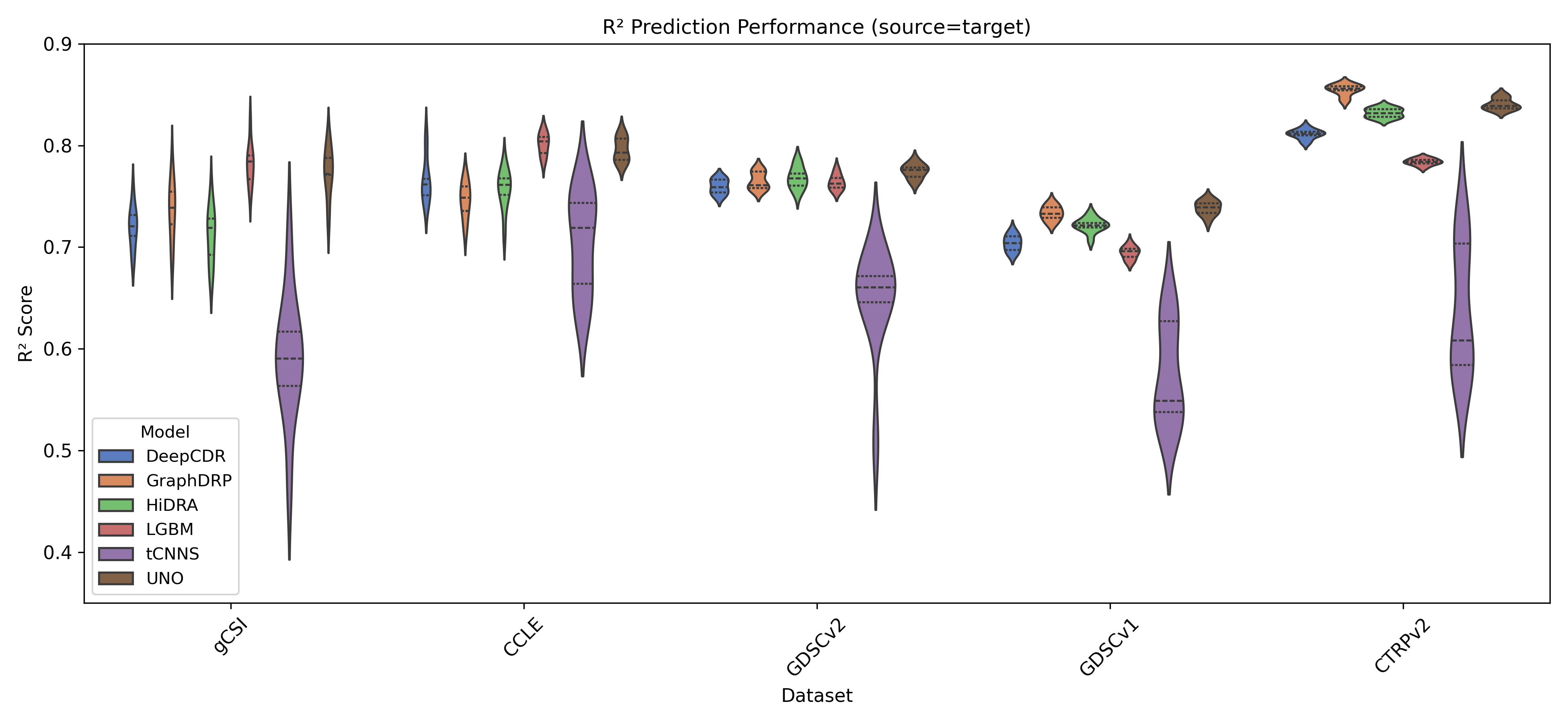}
    \caption{Violin plot showing \( R^2 \) performance scores across models, datasets, and data splits in the within-dataset analysis.
    Each violin plot shows the approximate probability density of the data, with the width indicating frequency of data points at each \( R^2 \) value. The markers in each violin plot indicate the median and interquartile range.
    Datasets are arranged by total sample size, representing the number of cell-drug response samples.
    }
    \label{fig:plot_within_dataset}
\end{figure}

Among the models, UNO achieves the highest mean \( R^2 \) computed across datasets (\( 0.785 \)), closely followed by GraphDRP (\( 0.767 \)), LGBM (\( 0.765 \)), HiDRA (\( 0.758 \)), and DeepCDR (\( 0.752 \)) (see Table \ref{tab:within_dataset_mean_r2} and Fig. \ref{fig:plot_within_dataset}). While UNO and GraphDRP perform slightly better than other models on the largest dataset, CTRPv2, differences in predictive accuracy between these top models are modest, and their performance ranges overlap when standard deviations are considered. LGBM, the only non-DL model in this analysis, achieves slightly higher predictive accuracy on smaller datasets (e.g., gCSI and CCLE), although the differences from DL models are not substantial. Conversely, tCNNS clearly exhibits lower mean accuracy (\( 0.632 \)) and higher variability, distinguishing it from the other models.

Model stability, measured by the standard deviation of \( R^2 \) scores across splits, reveals notable differences among models. LGBM demonstrates the lowest variability (\( 0.0096 \)), reflecting stable performance across splits (see Table \ref{tab:within_dataset_std_r2}). It is followed by UNO, DeepCDR, GraphDRP, and HiDRA. In contrast, tCNNS exhibits the highest variability (\( 0.0548 \)) across all datasets, indicating inconsistent performance across data splits.

Among the datasets, CTRPv2 and GDSCv1 exhibit contrasting patterns. CTRPv2 achieves the highest mean \( R^2 \) computed across models (\( 0.794 \)) and notably low prediction variability, suggesting robust data patterns that facilitate accurate and stable predictions. In contrast, GDSCv1 shows lower \( R^2 \) scores across models (\( 0.694 \)), highlighting challenges in capturing strong associations between input features and response data. These findings position CTRPv2 as a relatively stable and predictable dataset for model evaluation, while GDSCv1 presents certain difficulties for obtaining accurate predictions.

The combination of mean \( R^2 \) and computing standard deviation provides complementary perspectives on model performance. High mean \( R^2 \) scores reflect a model's ability to achieve strong predictive accuracy, while low standard deviation indicates stable performance across splits. For example, LGBM combines relatively high predictive performance with good stability, making it a reliable choice for smaller datasets. Conversely, tCNNS underperforms in both predictive accuracy and stability, limiting its utility for within-dataset applications.

These within-dataset results highlight the varying strengths and weaknesses of different models and datasets, providing a foundation for understanding broader cross-dataset trends.

\begin{table}[ht]
    \centering
    \caption{\textbf{Within-dataset analysis (mean of \( R^2 \) scores)}. Each value represents the mean \( R^2 \) computed across data splits for a given model-dataset pair. Each row corresponds to the diagonal entries in \( G \) matrices, shown in Fig. \ref{fig:G_and_Gn_matrices}. In addition, \textit{Mean across datasets} and \textit{Mean across models} summarize the average performance across datasets and models, respectively. The datasets are ordered by increasing sample size, from gCSI to CTRPv2.
    }
    \label{tab:within_dataset_mean_r2}
    \begin{tabular}{lcccccc}
        \toprule
        Model    & gCSI  & CCLE  & GDSCv2 & GDSCv1 & CTRPv2 & Mean across datasets \\
        \hline
        DeepCDR  & 0.720 & 0.766 & 0.760  & 0.704  & 0.811  & 0.7522 \\
        GraphDRP & 0.736 & 0.746 & 0.765  & 0.733  & 0.855  & 0.7670 \\
        HiDRA    & 0.711 & 0.756 & 0.768  & 0.722  & 0.832  & 0.7578 \\
        LGBM     & 0.782 & 0.801 & 0.764  & 0.695  & 0.784  & 0.7652 \\
        tCNNS    & 0.591 & 0.705 & 0.648  & 0.575  & 0.639  & 0.6316 \\
        UNO      & 0.774 & 0.796 & 0.775  & 0.738  & 0.841  & 0.7848 \\
        \hline
        Mean across models & 0.719 & 0.762 & 0.747 & 0.694 & 0.794 & \\
        \bottomrule
    \end{tabular}
\end{table}

\begin{table}[ht]
    \centering
    \caption{\textbf{Within-dataset analysis (Standard deviation of \( R^2 \) scores)}. Each value represents the standard deviation of \( R^2 \) computed across data splits for a given model-dataset pair. Each row corresponds to the diagonal entries in parentheses in \( G \) matrices, shown in Fig. \ref{fig:G_and_Gn_matrices}. In addition, \textit{Mean across datasets} and \textit{Mean across models} provide aggregated variability measures across datasets and models, respectively. The datasets are ordered by increasing sample size, from gCSI to CTRPv2.
    }
    \label{tab:within_dataset_std_r2}
    \begin{tabular}{lcccccc}
        \toprule
        Model    & gCSI  & CCLE  & GDSCv2 & GDSCv1 & CTRPv2 & Mean across datasets \\
        \hline
        DeepCDR  & 0.020 & 0.023 & 0.007  & 0.008  & 0.005  & 0.0126 \\
        GraphDRP & 0.029 & 0.018 & 0.008  & 0.007  & 0.006  & 0.0136 \\
        HiDRA    & 0.027 & 0.020 & 0.011  & 0.007  & 0.005  & 0.0140 \\
        LGBM     & 0.020 & 0.011 & 0.008  & 0.006  & 0.003  & 0.0096 \\
        tCNNS    & 0.061 & 0.049 & 0.052  & 0.049  & 0.063  & 0.0548 \\
        UNO      & 0.025 & 0.012 & 0.007  & 0.007  & 0.006  & 0.0114 \\
        \hline
        Mean across models & 0.030 & 0.022 & 0.016 & 0.014 & 0.015 & \\
        \bottomrule
    \end{tabular}
\end{table}

\subsection{Cross-Dataset Performance}  \label{sec:cross_dataset_performance}

\begin{figure}[h]
    \centering
    \includegraphics[width=1.0\textwidth]{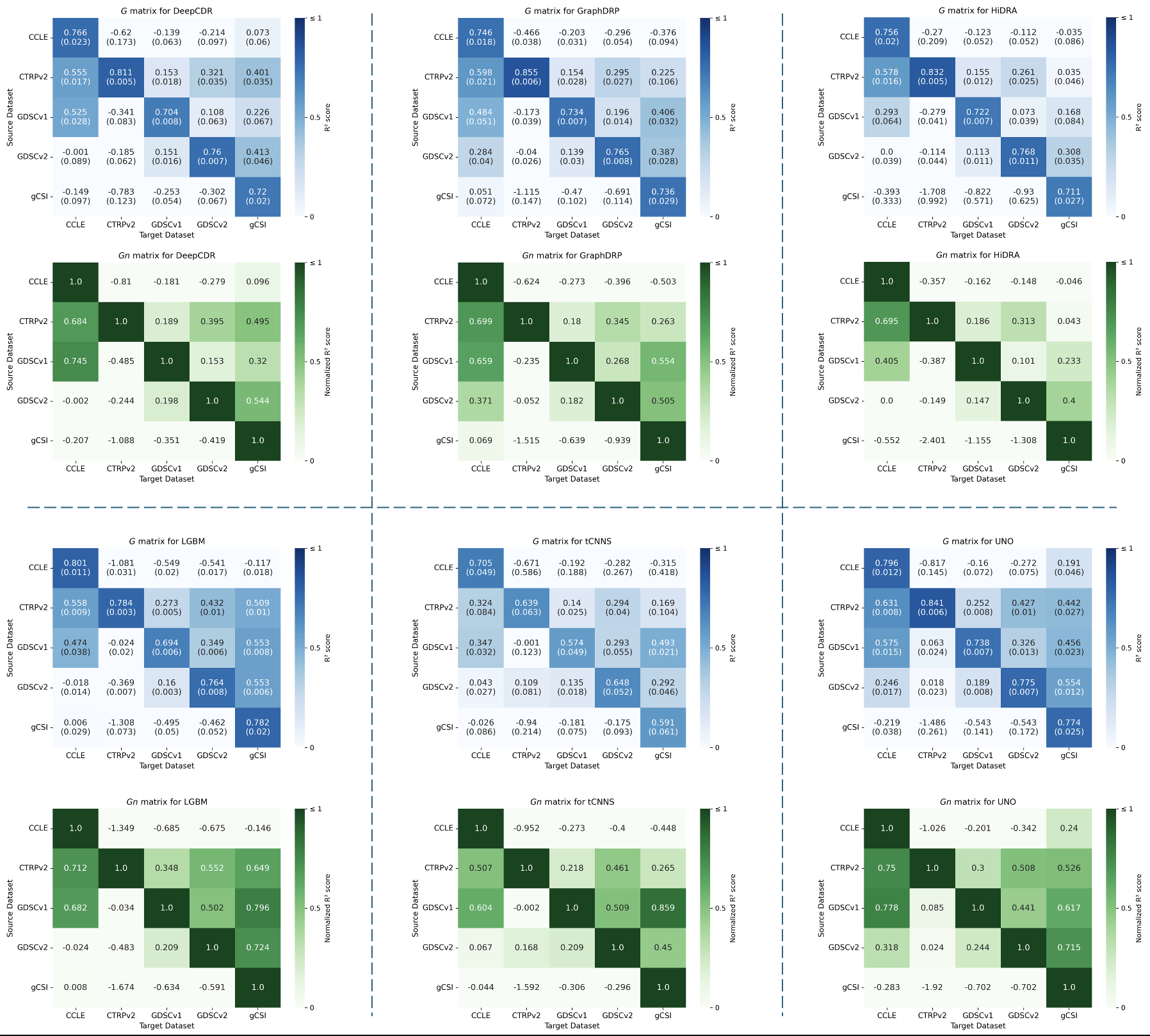}
    \caption{\textbf{Cross-dataset performance matrices for each model}. Each pair includes the basic matrix \( G \) \ref{sec:G_matrix} (blue) and the normalized matrix \( G_n \) \ref{sec:Gn_matrix}  (green). In the \( G \) matrices, values represent mean \( R^2 \) scores across splits, and the numbers in parentheses indicate standard deviations.}
    \label{fig:G_and_Gn_matrices}
\end{figure}

The cross-dataset evaluation underscores the inherent challenges in generalizing DRP models to unseen datasets. This can be observed by the more saturated blue shades on the matrix diagonal (representing within-dataset scores) as compared to the off-diagonal (representing cross-dataset scores) in the heatmaps of \( G \) for the different models, shown in Fig. \ref{fig:G_and_Gn_matrices}. Many entries in these matrices exhibit negative \( R^2 \) scores (light blue), indicating the inability of models to learn meaningful mappings between input features and treatment response. Despite the substantial differences between within- and cross-dataset scores, certain model-dataset pairs exhibit promising results.

\begin{figure}[h]
    \centering
    \includegraphics[width=1.0\textwidth]{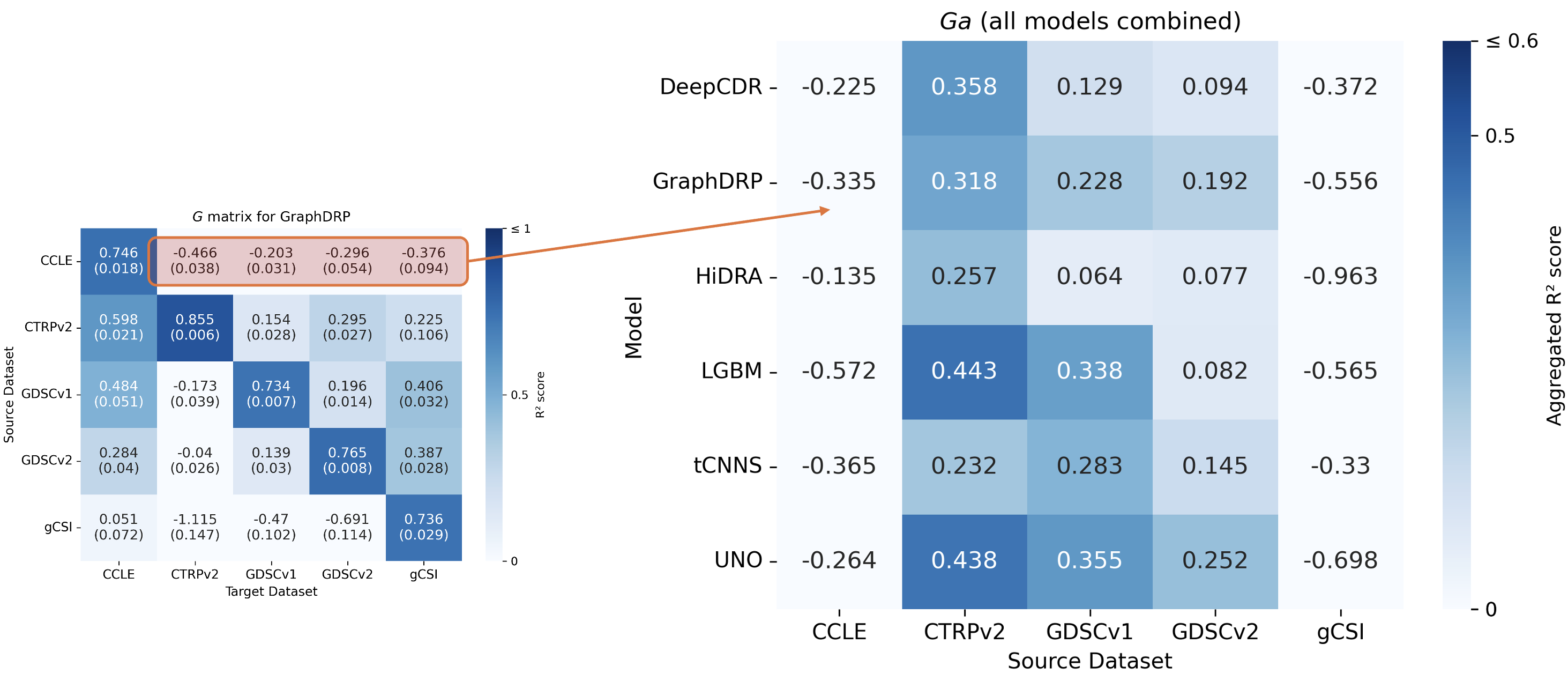}
    \caption{\( Ga \) matrix \ref{sec:Ga_matrix} for six drug response prediction models (five DL models and one based on LightGBM) and five drug response datasets \ref{sec:dataset}. An example average calculation is shown for a \(G_{a}[GraphDRP, CCLE] \) entry.
    }
    \label{fig:Ga_matrix}
\end{figure}

Models such as UNO and GraphDRP achieve relatively better \( R^2 \) scores in some cross-dataset scenarios compared to others (e.g., UNO: \( G[CTRPv2, CCLE] = 0.631 \), GraphDRP: \( G[CTRPv2, CCLE] = 0.598 \)), although performance differences among the top models remain modest. The aggregated cross-dataset metric \( G_a \) (Fig. \ref{fig:Ga_matrix}) indicates a slight advantage for UNO and GraphDRP when trained on larger datasets (CTRPv2, GDSCv1, GDSCv2). LGBM also demonstrates relatively strong generalization on certain dataset pairs, particularly CTRPv2 and GDSCv1, though it shows somewhat reduced performance on GDSCv2. By contrast, tCNNS, HiDRA, and DeepCDR underperform on average compared to others.

\begin{figure}[h]
    \centering
    \includegraphics[width=1.0\textwidth]{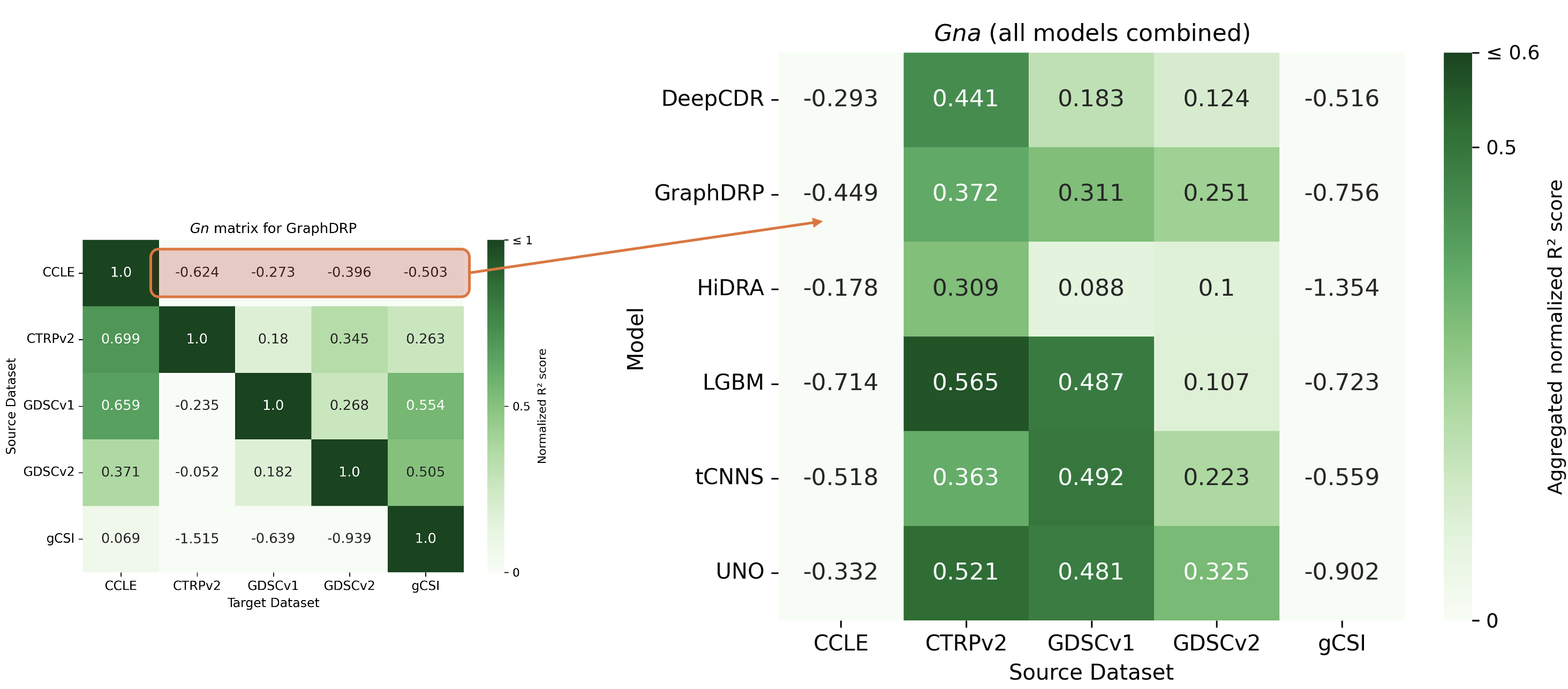}
    \caption{\( Gna \) matrix \ref{sec:Gna_matrix} for six drug response prediction models (five DL models and one based on LightGBM) and five drug response datasets \ref{sec:dataset}. An example average calculation is shown for a \(G_{na}[GraphDRP, CCLE] \) entry.
    }
    \label{fig:Gna_matrix}
\end{figure}

The benefits of normalized metrics \( G_n \) and \( G_{na} \) become evident when examining relative generalization performance.
While HiDRA performs well in the within-dataset setting, its performance substantially drops in the cross-dataset evaluations. This decline is better captured by \( G_n \) and \( G_{na} \), which contextualize generalization relative to within-dataset performance (see Fig. \ref{fig:G_and_Gn_matrices} and Fig. \ref{fig:Gna_matrix}). For example, HiDRA exhibits a sharp performance drop when transitioning from within-dataset to cross-dataset evaluation, with \( G_{na}[HiDRA, GDSCv1] = 0.088 \).
In contrast, UNO exhibits consistently good generalization in both within- and cross-dataset settings.
Notably, a high \( G_{na} \) value does not always indicate strong generalization performance, as observed with tCNNS. Although tCNNS exhibits relatively high \( G_{na} \), this is largely due to its already low within-dataset performance, resulting in a smaller relative decline when transitioning across datasets. This underscores the complementary nature of \( G_a \) and \( G_{na} \), as \( G_a \) captures absolute generalization performance, while \( G_{na} \) highlights relative performance shifts.

\begin{figure}[h]
    \centering
    \includegraphics[width=1.0\textwidth]{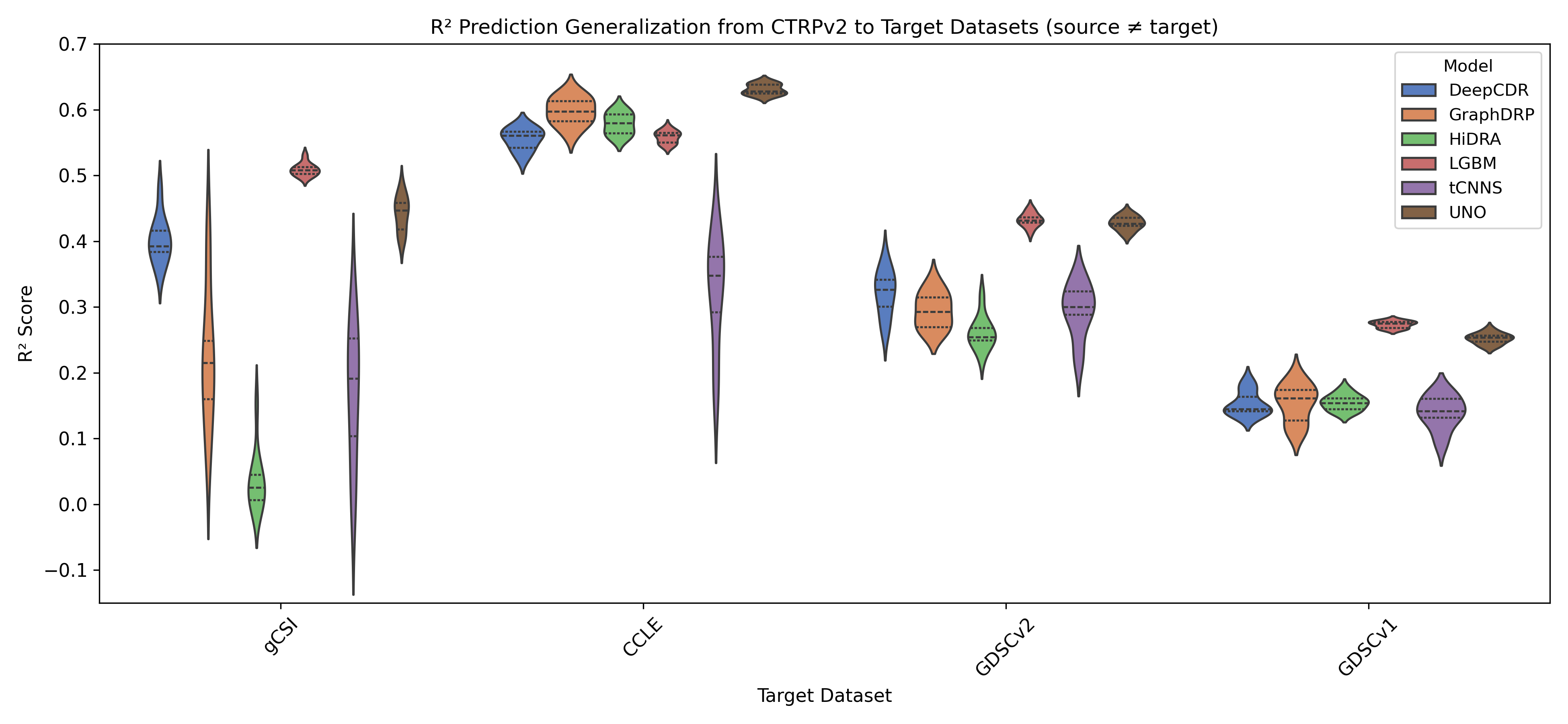}
    \caption{Violin plot showing model generalization performance (\( R^2 \) scores) when trained on CTRPv2 and evaluated on target datasets (gCSI, CCLE, GDSCv2, and GDSCv1) in the cross-dataset analysis.
    Each violin plot shows the approximate probability density of the data, with the width indicating frequency of data points at each \( R^2 \) value. The markers in each violin plot indicate the median and interquartile range.
    Datasets are ordered by total sample size, representing the number of cell-drug response samples.
    }
    \label{fig:plot_cross_dataset_from_CTRPv2}
\end{figure}

Among the datasets, CTRPv2 emerges as the most effective training source, producing higher generalization scores, highlighting the value of large, high-quality datasets for improving model transferability. The violin plots in Fig. \ref{fig:plot_cross_dataset_from_CTRPv2} provide a closer look at CTRPv2 as a source dataset, showing its generalization capability to the remaining target datasets.
Conversely, gCSI and CCLE consistently result in poor cross-dataset generalization, with mostly negative and close-to-zero \( R^2 \) scores, highlighting inherent challenges as a source dataset for generalization objectives.

These results highlight the varying degrees to which different models generalize across datasets and emphasize the importance of systematic evaluation frameworks. While direct statistical significance analyses are not reported due to the absence of comprehensive hyperparameter optimization (HPO), the observed trends remain informative for assessing model generalization. The proposed metrics (\( G \), \( G_a \), \( G_n \), and \( G_{na} \)) provide complementary insights into cross-dataset generalization, offering a more nuanced perspective on model transferability across datasets.

\section{Discussion} \label{sec:discussion}
In this work, we present a benchmarking framework for assessing the prediction generalization capabilities of DRP models. The framework incorporates five drug response datasets (Section \ref{sec:dataset}), six DRP models implemented within a unified code structure (Section \ref{sec:models}), software tools enabling scalable execution of the evaluation workflow (Section \ref{sec:csa}), and four metrics designed to assess the generalization capabilities of models (Section \ref{sec:metrics}). As part of the IMPROVE project \cite{weil15PB003IMPROVE2024}, this study underscores both the importance and challenges of rigorous benchmarking, with a particular emphasis on evaluating cross-dataset generalization, an essential but underexplored aspect of DRP model assessment.

Our key findings reveal substantial disparities between within- and cross-dataset performance. All models exhibit a significant drop in prediction accuracy when applied to external datasets, highlighting the persistent challenge of out-of-distribution (OOD) generalization.
However, certain models, such as UNO, GraphDRP, and LGBM, demonstrate moderately better performance within individual datasets and retain a portion of their predictive capability across datasets, although differences among these top models remain subtle.
% ...
From the dataset perspective, CTRPv2 emerges as the most effective source dataset, consistently producing higher generalization scores across multiple target datasets. This is likely due to its larger sample size and greater diversity in both unique cell lines and drug compounds, factors that contribute to better generalization.
Moreover, the proposed evaluation metrics (see Section \ref{sec:metrics}) provide complementary insights into both absolute and relative generalization capabilities. These metrics prove instrumental in identifying robust models while also highlighting those that exhibit the most pronounced decline in cross-dataset generalization relative to their within-dataset performance (see Fig. \ref{fig:Gna_matrix}).

Assessing generalization across drug response datasets is not a novel concept in DRP research. Many recent studies acknowledge the limitations of within-dataset evaluation and attempt to validate their models using a single external dataset. However, such validation strategy may not fully capture the complexities of model transferability across contexts. Furthermore, our prior work has already explored cross-dataset generalization through similar \( d \)-by-\( d \) evaluations \cite{xiaCrossstudyAnalysisDrug2022, partinAbstract5380Systematic2023}. Our current study builds upon these efforts by introducing a structured benchmarking framework designed to enable fair and scalable comparisons of models. By ensuring consistent data preprocessing, standardized evaluation metrics, and rigorous experimental design, this framework provides a more reliable means of assessing model robustness in cross-dataset scenarios. We emphasize that the goal of this work is not to demonstrate the superiority of a particular model but rather to present a framework that highlights the challenges associated with cross-dataset analysis. A structured evaluation methodology is critical for accurately assessing the reliability and transferability of AI-based DRP models into real-world biomedical applications.

The observed cross-dataset performance degradation may not solely stem from challenges inherent to OOD generalization, where the distributions of training and testing data differ, but also reflect dataset-specific experimental differences. For instance, as noted previously in Xia et al. \cite{xiaCrossstudyAnalysisDrug2022}, GDSCv1 utilizes a different viability assay (Syto60) compared to other datasets such as CTRPv2, gCSI, and CCLE (which primarily use the CellTiter-Glo). Such methodological differences in experimental protocols likely contribute to systematic variability across datasets, further complicating model generalization.


This study has several important limitations.
First, the datasets were obtained from different studies, each potentially contributing unique biases stemming from varying experimental conditions and assay protocols. Differences in data generation and collection can impact model performance and adversely impact generalizability.
Second, while the standardized code for training and inference follows a similar pattern across models, the preprocessing stage remains highly model-dependent. Establishing more structured guidelines for preprocessing would enhance standardization and improve the reliability of benchmarking.
Third, we employed random data partitioning to obtain the data splits, which can lead to inflated performance due to potential data leakage across cell line and drugs. Future work should explore more stringent splitting strategies, such as drug-blind and cell-blind splits, to better simulate real-world scenarios \cite{partinDeepLearningMethods2023}.
Finally, the reported results were obtained using limited HPO. While our primary objective is to examine broad trends and generalization patterns rather than optimizing predictive accuracy, a more exhaustive HPO search could refine model performance. Consequently, while the results highlight performance trends, definitive statements about the superiority of specific models should be interpreted cautiously.





Future work could build upon this benchmarking framework by exploring learning techniques such as transfer learning \cite{prassePreTrainingVitroFineTuning2022, sharifi-noghabiAITLAdversarialInductive2020, peresdasilvaTUGDATaskUncertainty2021, zhuEnsembleTransferLearning2020, maFewshotLearningCreates2021} to mitigate the impact of OOD predictions, thereby improving cross-dataset performance and narrowing the gap between within- and cross-dataset generalization. Given the modular design of the existing framework, incorporating transfer learning methods would require moderate modifications, leveraging the resources described in Section \ref{sec:methods}.
Another promising direction involves extending the evaluation scheme beyond single-valued generalization metrics to alternative assessments, such as studying how model performance scales with increasing data availability \cite{partinLearningCurvesDrug2021, bransonUnderstandingSourcesPerformance2024}, comparing model interpretability for clinical decision-making \cite{liInterpretableDeepLearning2023}, or integrating evaluation methodologies relevant to drug discovery applications \cite{guoUMAPclusteringSplitRigorous2024, narykovIntegrationComputationalDocking2024}.
Additionally, data-centric approaches have proven effective in enhancing prediction accuracy in traditional ML benchmarks, making them particularly valuable in biomedical applications where data scarcity is a major challenge. A key future direction is to systematically construct training and evaluation sets to align more closely with the data distribution of target datasets, ensuring that models are better suitable for real-world deployment in preclinical and clinical settings.


In conclusion, this study highlights the pressing need for rigorous benchmarking in DRP model development. Rather than identifying a single superior model, our framework enables fair and meaningful model comparisons, emphasizing the necessity of structured evaluation methodologies that reflect real-world complexities in predictive oncology.
Scientists across various domains continue to adapt advanced AI techniques for diverse applications, yet without rigorous benchmarking practices, it remains challenging to assess the true potential and limitations of these approaches \cite{malakarBenchmarkingMachineLearning2018, farjonAgroCountersRepositoryCounting2024, blizerComparingMLMethods2024, utyamishevNetwiseDetectionHardware2024}. 
Robust benchmarking frameworks are indispensable for assessing the true potential of prediction models, cutting through hype, fostering trust, and bridging the crucial gap between AI-based DRP models and their adoption in preclinical and clinical applications.

\section*{Data and Code Availability}

\begin{enumerate}

    \item \textbf{Reproduce}. The code and data required to reproduce the plots and tables in this paper are available in a GitHub repository \href{https://github.com/adpartin/cross-dataset-drp-paper/}{https://github.com/adpartin/cross-dataset-drp-paper/}.

    \item \textbf{Data}. The benchmark data, including drug response data, omics and drug representations, and data splits, are available in this \href{http://web.cels.anl.gov/projects/IMPROVE_FTP/candle/public/improve/benchmarks/single_drug_drp/benchmark-data-pilot1/csa_data/}{link}.

    \item \textbf{Models}. The standardized models used in this study are available in these links:
\href{https://github.com/JDACS4C-IMPROVE/DeepCDR}{DeepCDR},
\href{https://github.com/JDACS4C-IMPROVE/HiDRA}{HiDRA},
\href{https://github.com/JDACS4C-IMPROVE/GraphDRP}{GraphDRP},
\href{https://github.com/JDACS4C-IMPROVE/LGBM}{LGBM},
\href{https://github.com/JDACS4C-IMPROVE/tCNNS-Project}{tCNNS},
\href{https://github.com/JDACS4C-IMPROVE/UNO}{UNO}.

    \item \textbf{Workflow}. The scalable cross-dataset analysis workflow is available in the IMPROVE \href{https://github.com/JDACS4C-IMPROVE/IMPROVE/tree/develop/workflows/csa/parsl}{repo}.
\end{enumerate}
\section*{Funding}
The research was part of the IMPROVE project under the NCI-DOE Collaboration Program that has been funded in whole or in part with Federal funds from the National Cancer Institute, National Institutes of Health, Task Order No. 75N91019F00134 and from the Department of Energy under Award Number ACO22002-001-00000.

\bibliographystyle{ieeetr}

\bibliography{zotero_03-09-2025}

\begin{thebibliography}{10}

\bibitem{singhalHallmarksPredictiveOncology2025}
A.~Singhal, X.~Zhao, P.~Wall, E.~So, G.~Calderini, A.~Partin, N.~Koussa, P.~Vasanthakumari, O.~Narykov, Y.~Zhu, S.~E. Jones, F.~{Abbas-Aghababazadeh}, S.~Kadambat~Nair, J.-C. {B{\'e}lisle-Pipon}, A.~Jayaram, B.~A. Parker, K.~T. Yeung, J.~I. Griffiths, R.~Weil, A.~Nath, B.~{Haibe-Kains}, and T.~Ideker, ``The {{Hallmarks}} of {{Predictive Oncology}},'' {\em Cancer Discovery}, pp.~OF1--OF15, Jan. 2025.

\bibitem{senftPrecisionOncologyRoad2017}
D.~Senft, M.~D.~M. Leiserson, E.~Ruppin, and Z.~A. Ronai, ``Precision {{Oncology}}: {{The Road Ahead}},'' {\em Trends in Molecular Medicine}, vol.~23, pp.~874--898, Oct. 2017.

\bibitem{partinDeepLearningMethods2023}
A.~Partin, T.~S. Brettin, Y.~Zhu, O.~Narykov, A.~Clyde, J.~Overbeek, and R.~L. Stevens, ``Deep learning methods for drug response prediction in cancer: {{Predominant}} and emerging trends,'' {\em Frontiers in Medicine}, vol.~10, 2023.

\bibitem{ballesterArtificialIntelligenceDrug2021}
P.~J. Ballester, R.~Stevens, B.~{Haibe-Kains}, R.~S. Huang, and T.~Aittokallio, ``Artificial intelligence for drug response prediction in disease models,'' {\em Briefings in Bioinformatics}, Oct. 2021.

\bibitem{adamMachineLearningApproaches2020}
G.~Adam, L.~Ramp{\'a}{\v s}ek, Z.~Safikhani, P.~Smirnov, B.~{Haibe-Kains}, and A.~Goldenberg, ``Machine learning approaches to drug response prediction: Challenges and recent progress,'' {\em npj Precision Oncology}, vol.~4, no.~1, 2020.

\bibitem{sharifi-noghabiDrugSensitivityPrediction2021}
H.~{Sharifi-Noghabi}, S.~{Jahangiri-Tazehkand}, P.~Smirnov, C.~Hon, A.~Mammoliti, S.~K. Nair, A.~S. Mer, M.~Ester, and B.~{Haibe-Kains}, ``Drug sensitivity prediction from cell line-based pharmacogenomics data: Guidelines for developing machine learning models,'' {\em Briefings in Bioinformatics}, vol.~22, Nov. 2021.

\bibitem{weil15PB003IMPROVE2024}
M.~Weil, A.~Partin, N.~Kousa, S.~Jones, S.~Gosline, T.~Brettin, and T.~I. {team;}, ``15 ({{PB003}}): {{The IMPROVE}} framework for automating and standardizing tumor drug response prediction modeling,'' {\em European Journal of Cancer}, vol.~211, p.~114544, Oct. 2024.

\bibitem{overbeekAssessingReusabilityDeep2024}
J.~C. Overbeek, A.~Partin, T.~S. Brettin, N.~Chia, O.~Narykov, P.~Vasanthakumari, A.~Wilke, Y.~Zhu, A.~Clyde, S.~Jones, R.~Gnanaolivu, Y.~Liu, J.~Jiang, C.~Wang, C.~Knutson, A.~McNaughton, N.~Kumar, G.~D. Fernando, S.~Ghosh, C.~{Sanchez-Villalobos}, R.~Zhang, R.~Pal, M.~R. Weil, and R.~L. Stevens, ``Assessing {{Reusability}} of {{Deep Learning-Based Monotherapy Drug Response Prediction Models Trained}} with {{Omics Data}},'' Sept. 2024.

\bibitem{keLightGBMHighlyEfficient2017}
G.~Ke, Q.~Meng, T.~Finley, T.~Wang, W.~Chen, W.~Ma, Q.~Ye, and T.-Y. Liu, ``{{LightGBM}}: {{A Highly Efficient Gradient Boosting Decision Tree}},'' in {\em Advances in {{Neural Information Processing Systems}}}, vol.~30, 2017.

\bibitem{zhuEnsembleTransferLearning2020}
Y.~Zhu, T.~Brettin, Y.~A. Evrard, A.~Partin, F.~Xia, M.~Shukla, H.~Yoo, J.~H. Doroshow, and R.~L. Stevens, ``Ensemble transfer learning for the prediction of anti-cancer drug response,'' {\em Scientific Reports}, vol.~10, no.~1, pp.~1--11, 2020.

\bibitem{parkSuperFELTSupervisedFeature2021}
S.~Park, J.~Soh, and H.~Lee, ``Super.{{FELT}}: Supervised feature extraction learning using triplet loss for drug response prediction with multi-omics data,'' {\em BMC Bioinformatics}, vol.~22, p.~269, May 2021.

\bibitem{xiaCrossstudyAnalysisDrug2022}
F.~Xia, J.~Allen, P.~Balaprakash, T.~Brettin, C.~{Garcia-Cardona}, A.~Clyde, J.~Cohn, J.~Doroshow, X.~Duan, V.~Dubinkina, Y.~Evrard, Y.~J. Fan, J.~Gans, S.~He, P.~Lu, S.~Maslov, A.~Partin, M.~Shukla, E.~Stahlberg, J.~M. Wozniak, H.~Yoo, G.~Zaki, Y.~Zhu, and R.~Stevens, ``A cross-study analysis of drug response prediction in cancer cell lines,'' {\em Briefings in Bioinformatics}, vol.~23, p.~bbab356, Jan. 2022.

\bibitem{sharifi-noghabiAITLAdversarialInductive2020}
H.~{Sharifi-Noghabi}, S.~Peng, O.~Zolotareva, C.~C. Collins, and M.~Ester, ``{{AITL}}: {{Adversarial Inductive Transfer Learning}} with input and output space adaptation for pharmacogenomics,'' {\em Bioinformatics}, vol.~36, pp.~i380--i388, 2020.

\bibitem{maFewshotLearningCreates2021}
J.~Ma, S.~H. Fong, Y.~Luo, C.~J. Bakkenist, J.~P. Shen, S.~Mourragui, L.~F.~A. Wessels, M.~Hafner, R.~Sharan, J.~Peng, and T.~Ideker, ``Few-shot learning creates predictive models of drug response that translate from high-throughput screens to individual patients,'' {\em Nature Cancer}, vol.~2, no.~2, pp.~233--244, 2021.

\bibitem{peresdasilvaTUGDATaskUncertainty2021}
R.~{Peres da Silva}, C.~Suphavilai, and N.~Nagarajan, ``{{TUGDA}}: Task uncertainty guided domain adaptation for robust generalization of cancer drug response prediction from in vitro to in vivo settings,'' {\em Bioinformatics}, vol.~37, pp.~i76--i83, 2021.

\bibitem{sharifi-noghabiOutofdistributionGeneralizationLabelled2021}
H.~{Sharifi-Noghabi}, P.~A. Harjandi, O.~Zolotareva, C.~C. Collins, and M.~Ester, ``Out-of-distribution generalization from labelled and unlabelled gene expression data for drug response prediction,'' {\em Nature Machine Intelligence}, vol.~3, pp.~962--972, Nov. 2021.

\bibitem{liuImprovingPredictionPhenotypic2019}
P.~Liu, H.~Li, S.~Li, and K.-S. Leung, ``Improving prediction of phenotypic drug response on cancer cell lines using deep convolutional network,'' {\em BMC Bioinformatics}, vol.~20, no.~408, 2019.

\bibitem{partinLearningCurvesDrug2021}
A.~Partin, T.~Brettin, Y.~A. Evrard, Y.~Zhu, H.~Yoo, F.~Xia, S.~Jiang, A.~Clyde, M.~Shukla, M.~Fonstein, J.~H. Doroshow, and R.~L. Stevens, ``Learning curves for drug response prediction in cancer cell lines,'' {\em BMC Bioinformatics}, vol.~22, May 2021.

\bibitem{bransonComparisonMultipleModalities2024}
N.~Branson, P.~R. Cutillas, and C.~Bessant, ``Comparison of multiple modalities for drug response prediction with learning curves using neural networks and {{XGBoost}},'' {\em Bioinformatics Advances}, vol.~4, p.~vbad190, Jan. 2024.

\bibitem{dengPathwayGuidedDeepNeural2020}
L.~Deng, Y.~Cai, W.~Zhang, W.~Yang, B.~Gao, and H.~Liu, ``Pathway-{{Guided Deep Neural Network}} toward {{Interpretable}} and {{Predictive Modeling}} of {{Drug Sensitivity}},'' {\em Journal of Chemical Information and Modeling}, vol.~60, pp.~4497--4505, Oct. 2020.

\bibitem{liInterpretableDeepLearning2023}
Y.~Li, D.~E. Hostallero, and A.~Emad, ``Interpretable deep learning architectures for improving drug response prediction performance: Myth or reality?,'' {\em Bioinformatics}, vol.~39, p.~btad390, June 2023.

\bibitem{barretinaCancerCellLine2012}
J.~Barretina, G.~Caponigro, N.~Stransky, K.~Venkatesan, A.~A. Margolin, S.~Kim, C.~J. Wilson, J.~Leh{\'a}r, G.~V. Kryukov, D.~Sonkin, A.~Reddy, M.~Liu, L.~Murray, M.~F. Berger, J.~E. Monahan, P.~Morais, J.~Meltzer, A.~Korejwa, J.~{Jan{\'e}-Valbuena}, F.~A. Mapa, J.~Thibault, E.~{Bric-Furlong}, P.~Raman, A.~Shipway, I.~H. Engels, J.~Cheng, G.~K. Yu, J.~Yu, P.~Aspesi, M.~De~Silva, K.~Jagtap, M.~D. Jones, L.~Wang, C.~Hatton, E.~Palescandolo, S.~Gupta, S.~Mahan, C.~Sougnez, R.~C. Onofrio, T.~Liefeld, L.~MacConaill, W.~Winckler, M.~Reich, N.~Li, J.~P. Mesirov, S.~B. Gabriel, G.~Getz, K.~Ardlie, V.~Chan, V.~E. Myer, B.~L. Weber, J.~Porter, M.~Warmuth, P.~Finan, J.~L. Harris, M.~Meyerson, T.~R. Golub, M.~P. Morrissey, W.~R. Sellers, R.~Schlegel, and L.~A. Garraway, ``The {{Cancer Cell Line Encyclopedia}} enables predictive modelling of anticancer drug sensitivity,'' {\em Nature}, vol.~483, pp.~603--607, 2012.

\bibitem{seashore-ludlowHarnessingConnectivityLargescale2015}
B.~{Seashore-Ludlow}, M.~G. Rees, J.~H. Cheah, M.~Coko, E.~V. Price, M.~E. Coletti, V.~Jones, N.~E. Bodycombe, C.~K. Soule, J.~Gould, B.~Alexander, A.~Li, P.~Montgomery, M.~J. Wawer, N.~Kuru, J.~D. Kotz, C.~{Suk-Yee Hon}, B.~Munoz, T.~Liefeld, V.~Dan{\v c}ik, J.~A. Bittker, M.~Palmer, J.~E. Bradner, A.~F. Shamji, P.~A. Clemons, and S.~L. Schreiber, ``Harnessing connectivity in a large-scale small-molecule sensitivity dataset,'' {\em Cancer Discovery}, vol.~5, no.~11, pp.~1210--1223, 2015.

\bibitem{havertyReproduciblePharmacogenomicProfiling2016}
P.~M. Haverty, E.~Lin, J.~Tan, Y.~Yu, B.~Lam, S.~Lianoglou, R.~M. Neve, S.~Martin, J.~Settleman, R.~L. Yauch, and R.~Bourgon, ``Reproducible pharmacogenomic profiling of cancer cell line panels,'' {\em Nature}, vol.~533, no.~7603, pp.~333--337, 2016.

\bibitem{yangGenomicsDrugSensitivity2013}
W.~Yang, J.~Soares, P.~Greninger, E.~J. Edelman, H.~Lightfoot, S.~Forbes, N.~Bindal, D.~Beare, J.~A. Smith, I.~R. Thompson, S.~Ramaswamy, P.~A. Futreal, D.~A. Haber, M.~R. Stratton, C.~Benes, U.~McDermott, and M.~J. Garnett, ``Genomics of {{Drug Sensitivity}} in {{Cancer}} ({{GDSC}}): {{A}} resource for therapeutic biomarker discovery in cancer cells,'' {\em Nucleic Acids Research}, vol.~41, pp.~955--961, 2013.

\bibitem{tsherniakDefiningCancerDependency2017}
A.~Tsherniak, F.~Vazquez, P.~G. Montgomery, B.~A. Weir, G.~Kryukov, G.~S. Cowley, S.~Gill, W.~F. Harrington, S.~Pantel, J.~M. {Krill-Burger}, R.~M. Meyers, L.~Ali, A.~Goodale, Y.~Lee, G.~Jiang, J.~Hsiao, W.~F.~J. Gerath, S.~Howell, E.~Merkel, M.~Ghandi, L.~A. Garraway, D.~E. Root, T.~R. Golub, J.~S. Boehm, and W.~C. Hahn, ``Defining a {{Cancer Dependency Map}},'' {\em Cell}, vol.~170, pp.~564--576.e16, July 2017.

\bibitem{rdkit}
{RDKit: Open-source cheminformatics} Last accessed: January 2023.
\newblock \url{https://www.rdkit.org}.

\bibitem{moriwakiMordredMolecularDescriptor2018}
H.~Moriwaki, Y.-S. Tian, N.~Kawashita, and T.~Takagi, ``Mordred: A molecular descriptor calculator,'' {\em Journal of Cheminformatics}, vol.~10, no.~4, 2018.

\bibitem{liuDeepCDRHybridGraph2020}
Q.~Liu, Z.~Hu, R.~Jiang, and M.~Zhou, ``{{DeepCDR}}: A hybrid graph convolutional network for predicting cancer drug response,'' {\em Bioinformatics}, vol.~36, no.~26, pp.~i911--i918, 2020.

\bibitem{nguyenGraphConvolutionalNetworks2021}
T.-T. Nguyen, G.~T.~T. Nguyen, T.~Nguyen, and D.-H. Le, ``Graph convolutional networks for drug response prediction,'' {\em IEEE/ACM Transactions on Computational Biology and Bioinformatics}, 2021.

\bibitem{jinHiDRAHierarchicalNetwork2021}
I.~Jin and H.~Nam, ``{{HiDRA}}: {{Hierarchical Network}} for {{Drug Response Prediction}} with {{Attention}},'' {\em Journal of Chemical Information and Modeling}, vol.~61, pp.~3858--3867, Aug. 2021.

\bibitem{partinAbstract5380Systematic2023}
A.~Partin, T.~S. Brettin, Y.~Zhu, J.~Overbeek, O.~Narykov, P.~Vasanthakumari, A.~Clyde, S.~E. Jones, S.~R. Ganakammal, J.~M. Wozniak, A.~Wilke, J.~{Mohd-Yusof}, M.~R. Weil, A.~T. Pearson, and R.~L. Stevens, ``Abstract 5380: {{Systematic}} evaluation and comparison of drug response prediction models: A case study of prediction generalization across cell lines datasets,'' {\em Cancer Research}, vol.~83, p.~5380, Apr. 2023.

\bibitem{babujiParslPervasiveParallel2019}
Y.~Babuji, A.~Woodard, Z.~Li, D.~S. Katz, B.~Clifford, R.~Kumar, L.~Lacinski, R.~Chard, J.~M. Wozniak, I.~Foster, M.~Wilde, and K.~Chard, ``Parsl: {{Pervasive Parallel Programming}} in {{Python}},'' in {\em Proceedings of the 28th {{International Symposium}} on {{High-Performance Parallel}} and {{Distributed Computing}}}, {{HPDC}} '19, (New York, NY, USA), pp.~25--36, Association for Computing Machinery, June 2019.

\bibitem{prassePreTrainingVitroFineTuning2022}
P.~Prasse, P.~Iversen, M.~Lienhard, K.~Thedinga, R.~Herwig, and T.~Scheffer, ``Pre-{{Training}} on {{In Vitro}} and {{Fine-Tuning}} on {{Patient-Derived Data Improves Deep Neural Networks}} for {{Anti-Cancer Drug-Sensitivity Prediction}},'' {\em Cancers}, vol.~14, p.~3950, Jan. 2022.

\bibitem{bransonUnderstandingSourcesPerformance2024}
N.~Branson, P.~R. Cutillas, and C.~Besseant, ``Understanding the {{Sources}} of {{Performance}} in {{Deep Learning Drug Response Prediction Models}},'' June 2024.

\bibitem{guoUMAPclusteringSplitRigorous2024}
Q.~Guo, S.~Hernandez, and P.~Ballester, ``{{UMAP-clustering}} split for rigorous evaluation of {{AI}} models for virtual screening on cancer cell lines,'' Dec. 2024.

\bibitem{narykovIntegrationComputationalDocking2024}
O.~Narykov, Y.~Zhu, T.~Brettin, Y.~A. Evrard, A.~Partin, M.~Shukla, F.~Xia, A.~Clyde, P.~Vasanthakumari, J.~H. Doroshow, and R.~L. Stevens, ``Integration of {{Computational Docking}} into {{Anti-Cancer Drug Response Prediction Models}},'' {\em Cancers}, vol.~16, p.~50, Jan. 2024.

\bibitem{malakarBenchmarkingMachineLearning2018}
P.~Malakar, P.~Balaprakash, V.~Vishwanath, V.~Morozov, and K.~Kumaran, ``Benchmarking {{Machine Learning Methods}} for {{Performance Modeling}} of {{Scientific Applications}},'' in {\em International {{Workshop}} on {{Performance Modeling}}, {{Benchmarking}} and {{Simulation}} of {{High Performance Computer Systems}} ({{PMBS}})}, (Dallas, TX, USA), pp.~33--44, IEEE, Nov. 2018.

\bibitem{farjonAgroCountersRepositoryCounting2024}
G.~Farjon and Y.~Edan, ``{{AgroCounters}}---{{A}} repository for counting objects in images in the agricultural domain by using deep-learning algorithms: {{Framework}} and evaluation,'' {\em Computers and Electronics in Agriculture}, vol.~222, p.~108988, July 2024.

\bibitem{blizerComparingMLMethods2024}
A.~Blizer, O.~Glickman, and I.~M. Lensky, ``Comparing {{ML Methods}} for {{Downscaling Near-Surface Air Temperature}} over the {{Eastern Mediterranean}},'' {\em Remote Sensing}, vol.~16, p.~1314, Jan. 2024.

\bibitem{utyamishevNetwiseDetectionHardware2024}
D.~Utyamishev and I.~{Partin-Vaisband}, ``Netwise {{Detection}} of {{Hardware Trojans Using Scalable Convolution}} of {{Graph Embedding Clouds}},'' {\em IEEE Transactions on Computer-Aided Design of Integrated Circuits and Systems}, vol.~43, pp.~3116--3128, Oct. 2024.

\end{thebibliography}

\end{document}